\documentclass[10pt,a4paper,twocolumn]{deepmind}

\usepackage{graphicx}
\usepackage{multirow}
\usepackage{amsmath,amssymb,amsfonts}
\usepackage{amsthm}
\usepackage{mathrsfs}
\usepackage[title]{appendix}
\usepackage{xcolor}
\usepackage{textcomp}
\usepackage{manyfoot}
\usepackage{booktabs}
\usepackage{booktabs}
\usepackage{multirow}
\usepackage{graphicx}
\usepackage[table]{xcolor} 
\usepackage{algorithm}
\usepackage{algorithmicx}
\usepackage{algpseudocode}
\usepackage{listings}
\usepackage{lipsum} 
\usepackage{tabularx}
\usepackage{hyperref}

\begin{document}


\title{Physics-Informed Spiking Neural Networks via Conservative Flux Quantization}

\author[1]{\fnm{Chi} \sur{Zhang}}
\email{chizhang23@mails.tsinghua.edu.cn (Work done during internship at NTU)}

\author*[2]{\fnm{Lin} \sur{Wang}}
\email{linwang@ntu.edu.sg}

\affil[1]{\orgdiv{Zhili College}, \orgname{Tsinghua University}, \orgaddress{\postcode{100084}, \country{China}}}

\affil[2]{\orgdiv{School of Electrical and Electronic Engineering}, \orgname{Nanyang Technological University}, \orgaddress{\postcode{639798}, \country{Singapore}}}

\abstract{%
Real-time, physically-consistent predictions on low-power edge devices is critical for the next generation embodied AI systems, yet it remains a major challenge. Physics-Informed Neural Networks (PINNs) 
combine data-driven learning with physics-based constraints to ensure the model's predictions are 
with underlying physical principles.
However, PINNs are energy-intensive and struggle to strictly enforce physical conservation laws. 
Brain-inspired spiking neural networks (SNNs) have emerged as a promising solution for edge computing and real-time processing. However, naively converting PINNs to SNNs degrades physical fidelity and fails to address long-term generalization issues. To this end, this paper introduce a novel Physics-Informed Spiking Neural Network (PISNN) framework. Importantly, to ensure strict physical conservation, we design the \textbf{Conservative Leaky Integrate-and-Fire (C-LIF) neuron}, whose dynamics structurally guarantee local mass preservation. To achieve robust temporal generalization, we introduce a novel \textbf{Conservative Flux Quantization (CFQ)} strategy, which redefines neural spikes as discrete packets of physical flux. Our CFQ learns a time-invariant physical evolution operator, enabling the PISNN to become a general-purpose solver -- \textbf{conservative-by-construction}. Extensive experiments show that our PISNN excels on diverse benchmarks. For both the canonical 1D heat equation and the more challenging 2D Laplace's Equation, it accurately simulates the system dynamics while maintaining perfect mass conservation by design -- a feat that is challenging for conventional PINNs. This work establishes a robust framework for fusing the rigor of scientific computing with the efficiency of neuromorphic engineering, paving the way for complex, long-term, and energy-efficient physics predictions for intelligent systems.
}

\maketitle


\begin{figure*}[t]
    \centering
    \includegraphics[width=.9\textwidth]{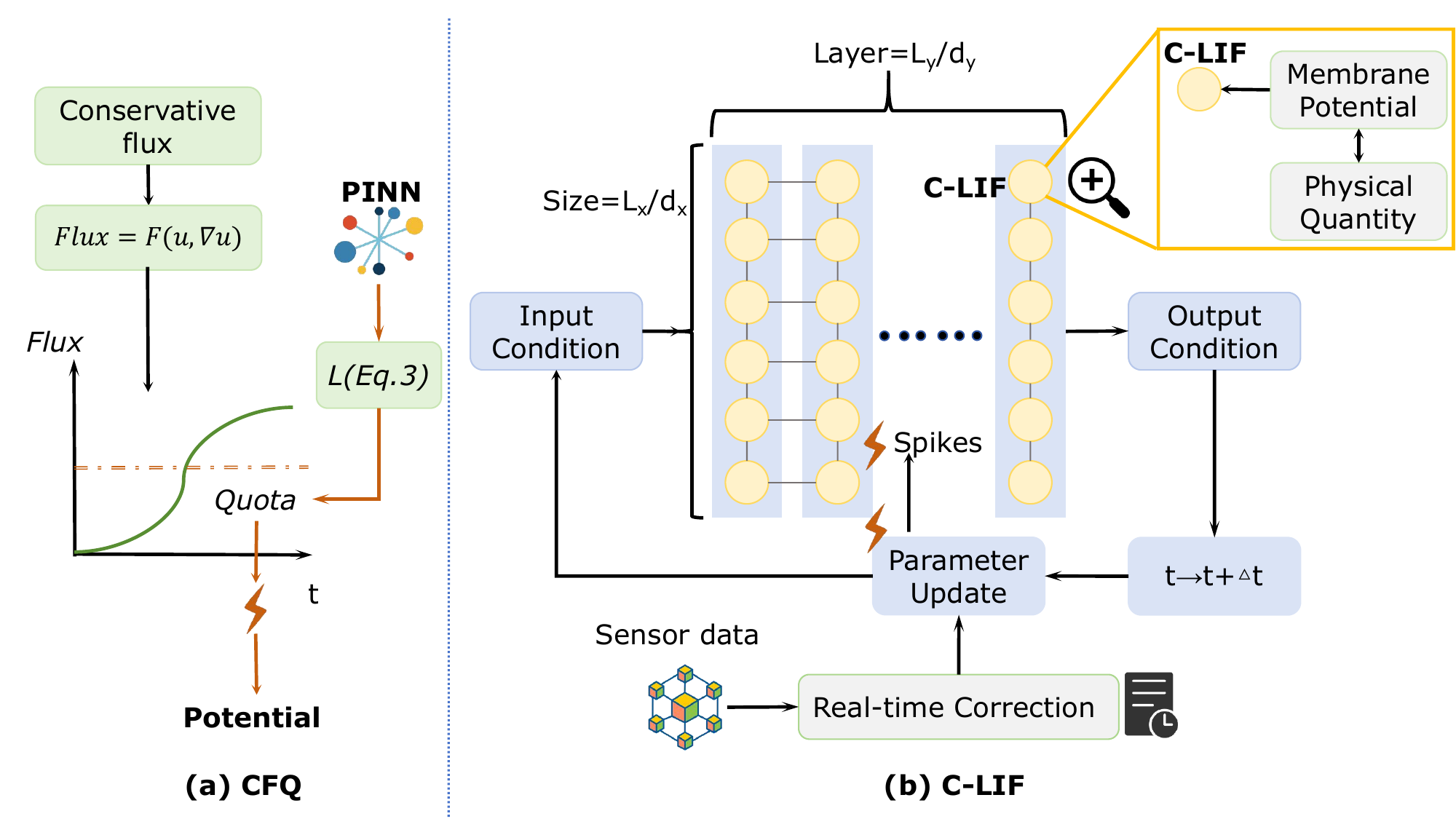}
    \vspace{-10pt}
\caption{\textbf{Schematic of the PISNN Framework.} 
\textbf{(a) Conservative Flux Quantization (CFQ).} Continuous physical fluxes are discretized into spike events via a learnable `quota' parameter. This threshold is optimized via distillation from a PINN teacher, bridging continuous physical laws with discrete neural signaling. 
\textbf{(b) Network Architecture and C-LIF Dynamics.} The network topology is strictly isomorphic to the physical grid, where spatial resolution defines network size ($L_x/d_x$). Each C-LIF neuron directly maps membrane potential to physical quantities, evolving through spike exchange. The architecture supports real-time parameter correction via external sensor data integration.}
    \label{fig:picture0}
\end{figure*}

\noindent
Deploying real-time, high-fidelity physics predictions on resource-constrained edge devices is a critical frontier for embodied AI~\cite{li2024behavior,newbury2024review}. For embodied agents such as drones and robots, the ability to perform on-device physics predictions is essential for safe training, online verification, and the generation of vast synthetic datasets for policy learning~\cite{ren2024infiniteworld,makoviychuk2021isaac,tobin2017domain}. However, a fundamental conflict exists between the escalating demand for physics prediction fidelity and the stringent power and computational limitations of edge hardware~\cite{peng2018sim,warden2019tinyml}. Migrating the predictions from power-intensive data centers is often unviable due to prohibitive network latency, creating an urgent need for a new class of numerical solvers that are simultaneously physically-consistent and energy-efficient~\cite{kokiadis2024decoupled}.

Traditional numerical solvers, such as finite difference or finite element methods, provide high precision and strong physical fidelity, but their cubic computational scaling renders them infeasible for real-time deployment on edge hardware with limited memory and energy budgets~\cite{lu2021priori,lahmer2022energy}. To overcome this barrier, learning-based approaches, especially Physics-Informed Neural Networks (PINNs), have emerged as flexible alternatives~\cite{raissi2017physics}. By embedding partial differential equation (PDE) residuals into the loss function, PINNs can approximate complex dynamics without explicit meshing. However, their reliance on dense matrix operations during inference results in considerable energy overhead, limiting their applicability to low-power environments~\cite{lahmer2022energy,wang2022respecting,hanafy2021design}. Moreover, PINNs often exhibit poor long-term generalization, as their predictive capability is constrained to the temporal domain seen during training~\cite{bonfanti2024generalization}. Recent advances in operator learning and reduced-order modeling have attempted to accelerate PDE solvers, yet they typically require large offline datasets and still face challenges in ensuring strict physical conservation. 

Spiking neural networks (SNNs) are brain-inspired neural models that simulate the neural dynamics with asynchronous spikes, in contrast to traditional artificial neural networks (ANNs) that process data synchronously \cite{furber2014spinnaker,zheng2023deep,liu2025neural}. This allows SNNs to operate in a more brain-like manner, and thus they have emerged as a promising solution for edge computing and real-time processing. 
However, naively converting PINNs to SNNs for energy efficiency purpose degrades physical fidelity and fails to address long-term generalization issues, as exemplified in a few preliminary efforts~\cite{zhang2023artificial,wang2022respecting,kapoor2024neural}.
This is because there lack conversion mechanisms explicitly designed around governing physical laws. Consequently, these methods tend to exacerbate the generalization limitations of PINNs, yield low physical fidelity, and remain less suitable for real-world deployment~\cite{mishra2022estimates,pfeiffer2018deep}. 

In this work, we introduce a novel solver architecture -- Physics-Informed Spiking Neural Network (\textbf{PISNN})-- that is inherently mass-conservative, as depicted in Fig.~\ref{fig:picture0}. The cornerstone of PISNN is an novel mechanism we term ‘\textbf{flux quantization}’, which forges a direct link between the local form of a physical conservation law and the event-driven dynamics of an SNN. This mechanism redefines the role of a neural spike, treating it as a discrete, quantized packet of physical flux, instead of an abstract signal. By ensuring that the system’s state can only evolve through the exchange of these flux packets, PISNN becomes ``\textbf{conservative-by-construction}'', embedding the physical law directly into its structure rather than approximating it via a loss function. This paper brings two major technical innovations. First, we introduce the Conservative Leaky Integrate-and-Fire (C-LIF) neuron that ensures the model's operation satisfies physical fidelity. Then, we propose a Conservative Flux Quantization (CFQ) method to guarantee the model's generalization. Finally, we achieve the deployment of this physics solver for edge devices.

In summary, our main contributions are three-fold. \textbf{First}, we formally establish the concept of flux quantization, providing a systematic methodology to map continuous physical dynamics onto the discrete computational fabric of SNNs. \textbf{Second}, by incorporating the continuity equation directly into the model’s structure, our PISNN achieves stable and accurate long-term forward evolution, breaking through a key limitation of existing learning-based solvers. \textbf{Third}, we demonstrate that this fusion of scientific computing principles and neuromorphic engineering provides a viable blueprint for deploying complex, energy-efficient physical simulations on edge devices, with significant implications for embodied AI and autonomous systems. 

In the following sections, we will detail the formulation of the PISNN, validate its performance on the 1D heat equation, and discuss its broader potential for on-device scientific computing.

\begin{table*}[t!]
\centering
\small 
\begin{tabularx}{\textwidth}{@{} l l c c c X @{}}
    \toprule
    \textbf{Solver Type} & \textbf{Core Mechanism} & \textbf{Conservation} & \textbf{Time Gen.} & \textbf{Cost} & \textbf{Primary Limitation} \\ \midrule
    
    \textbf{Standard PINN} \cite{raissi2017physics} & Continuous Mapping & Soft Constraint & Poor & High & Black-box physics; \\
    & (Loss Penalty) & (Approximate) & (Fails $t > T$) & (Dense FLOPs) & High training latency \cite{lahmer2022energy} \\ 
    \addlinespace[0.6em]
    
    \textbf{Neural Operator} \cite{mehtaj2025scientific} & Global Data Mapping & None & Good & Medium & Data-hungry; \\
    (FNO / DeepONet) & (Data-Driven) & (No Guarantee) & (In-Dist.) & (FFT Ops) & Non-conservative \cite{mehtaj2025scientific} \\ 
    \addlinespace[0.6em]
    
    \textbf{SNN} \cite{zhang2023artificial} & Rate-Coded Spiking & Violated & Unstable & \textbf{Low} & \textbf{Physical Distortion}; \\
    (Direct Conversion) & (Reset Mechanism) & (Reset Error) & (Accumulation) & (Spikes) & Info Loss \cite{zhang2023artificial} \\ 
    \addlinespace[0.6em]
    
    \textbf{PISNN (Ours)} & \textbf{Flux Quantization} & \textbf{Exact} & \textbf{Excellent} & \textbf{Very Low} & \textbf{None} \\ 
    & \textbf{(C-LIF Dynamics)} & \textbf{(Structural)} & \textbf{(Zero-Shot)} & \textbf{(Event-Driven)} & (Ideal for Edge) \\ 
    \bottomrule
\end{tabularx}

\caption{
\textbf{ Theoretical comparison revealing the deployment gap.} Current solvers fail to reconcile high physical fidelity with the strict resource constraints of edge devices. While traditional methods are computationally prohibitive \cite{lu2021priori}, data-driven approaches lack structural conservation \cite{raissi2017physics} or generalization capabilities \cite{bonfanti2024generalization}. PISNN uniquely fills this void for edge computing.}
\label{tab:picutre11}
\end{table*}
\section*{Results}
\begin{table*}[t!]
\centering
\small 
\begin{tabularx}{\textwidth}{@{} l l c c l @{}}
    \toprule
    \textbf{Task} & \textbf{Model Variant} & \textbf{Accuracy(RMSE)} & \textbf{Conservation(Mass Error)} & \textbf{Computational Cost(Solver Proxy)} \\ 
 \midrule

    \multirow{5}{*}{\textbf{1D Heat Eq.}} 
    & PINN~\cite{raissi2017physics} & $1.5 \times 10^{-4}$ & $3.5 \times 10^{-5}$ & $\sim 3.2 \times 10^{10}$ FLOPs \\ 
    & cPINN~\cite{jagtap2020conservative} & $\sim 1.2 \times 10^{-4}$ \textsuperscript{\dag} & $\sim 1.0 \times 10^{-5}$ \textsuperscript{\dag} & $\sim 4.8 \times 10^{10}$ FLOPs \textsuperscript{\ddag} \\
    & gPINN~\cite{yu2022gradient} & $\sim 1.0 \times 10^{-5}$ \textsuperscript{\dag} & $\sim 1.0 \times 10^{-5}$ \textsuperscript{\dag} & $\sim 1.6 \times 10^{11}$ FLOPs \textsuperscript{\ddag} \\
    & FNO~\cite{li2020fourier} & $\sim 10^{-5}$ \textsuperscript{\dag} & $\sim 10^{-3}$ (Poor) \textsuperscript{\dag} & $\sim 10^{9}$ FLOPs (FFT) \\
    
    \cmidrule{2-5} 
    
    & \textbf{PISNN (Ours)} & $3.1 \times 10^{-3}$ & $\mathbf{7.0 \times 10^{-7}}$ & $\mathbf{8.0 \times 10^{7}}$ \textbf{Spikes} \\ 
    \midrule
    
    \multirow{5}{*}{\textbf{2D Laplace}} 
    & PINN~\cite{raissi2017physics} & $2.0 \times 10^{-3}$ & $2.4 \times 10^{-4}$ & $\sim 2.7 \times 10^{12}$ FLOPs \\ 
    & cPINN~\cite{jagtap2020conservative} & $\sim 1.8 \times 10^{-3}$ \textsuperscript{\dag} & $\sim 10^{-5}$ \textsuperscript{\dag} & $\sim 4.0 \times 10^{12}$ FLOPs \textsuperscript{\ddag} \\
    & gPINN~\cite{yu2022gradient} & $\sim 2.0 \times 10^{-4}$ \textsuperscript{\dag} & $\sim 1.0 \times 10^{-5}$ \textsuperscript{\dag} & $\sim 1.4 \times 10^{13}$ FLOPs \textsuperscript{\ddag} \\
    & FNO~\cite{li2020fourier} & $\sim 5.0 \times 10^{-4}$ \textsuperscript{\dag} & $\sim 10^{-3}$ (Poor) \textsuperscript{\dag} & $\sim 10^{10}$ FLOPs (FFT) \\
    
    \cmidrule{2-5}
    
    & \textbf{PISNN (Ours)} & $\mathbf{1.39 \times 10^{-3}}$ & $\mathbf{3.5 \times 10^{-5}}$ & $\mathbf{1.2 \times 10^{10}}$ \textbf{Spikes} \\ 
    \bottomrule
\end{tabularx}

\vspace{0.1cm}

{\scriptsize 
\textsuperscript{\dag} Representative magnitude reported in original papers for similar benchmarks; exact values may vary with grid configuration. 
\textsuperscript{\ddag} Theoretical FLOPs estimated based on network architecture scaling.
}
\caption{
\textbf{Quantitative performance verification.} Compared to the PINN teacher and its variants, PISNN maintains competitive accuracy while achieving superior physical fidelity (reducing mass error by $\sim 10^2$ times) and drastically lower computational overhead (reducing cost metrics by $\sim 10^3$ times), confirming its viability for energy-constrained physics engines.}
\label{tab:master_comparison_final}
\end{table*}

We first compare PISNN against various physics solvers capable of edge deployment, as illustrated in Tab.~\ref{tab:picutre11}. We observe that mainstream solvers currently struggle with edge deployment due to multifaceted constraints, whereas our PISNN seamlessly addresses these challenges. The arguments supporting PISNN's perfect adaptability are detailed in the subsequent sections. Furthermore, we benchmark PISNN against PINN across diverse equations to validate its superior performance and low energy consumption. Results are shown in Tab.~\ref{tab:master_comparison_final}

\begin{figure*}[t]
    \centering
    \includegraphics[width=\textwidth]{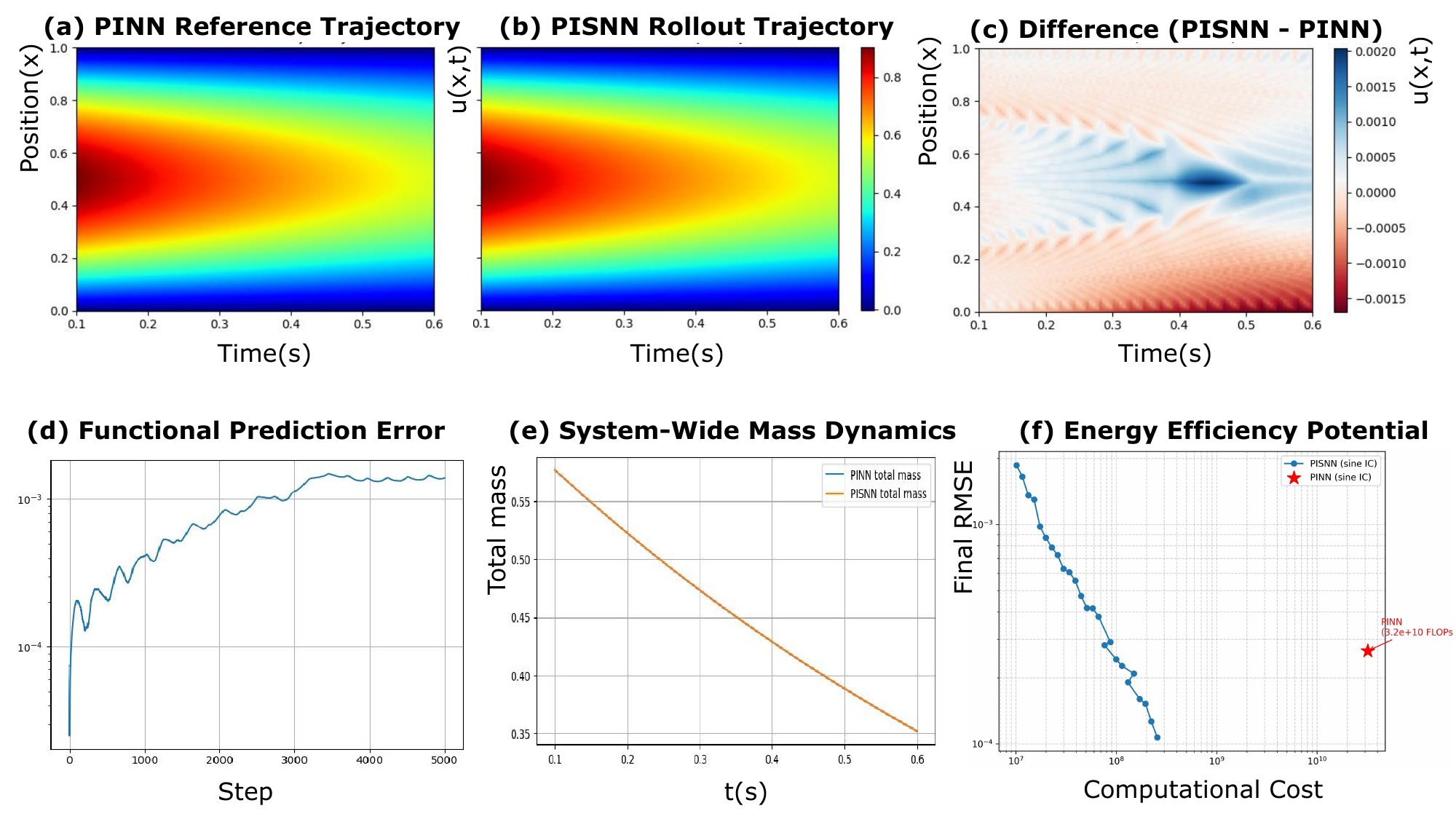}
    \vspace{-10pt}
\caption{\textbf{Comprehensive validation on the 1D heat equation showing high fidelity, conservation, and efficiency.} 
\textbf{(a-b)} Spatio-temporal evolution of the reference PINN and the zero-shot compiled PISNN. The PISNN rollout is visually indistinguishable from the teacher model.
\textbf{(c)} Point-wise error map $(u_{PISNN} - u_{PINN})$, showing negligible deviations.
\textbf{(d)} Temporal evolution of RMSE, stabilizing at the order of $10^{-4}$.
\textbf{(e)} System-wide mass dynamics. The PISNN trajectory (orange) perfectly tracks the reference (blue), confirming strict conservation adherence.
\textbf{(f)} Energy efficiency Pareto frontier. The PISNN (blue curve) allows a tunable trade-off between accuracy and spike count (quota). Notably, it matches PINN accuracy (red star) with a computational cost (spikes) approximately three orders of magnitude lower than the PINN's FLOPs.}
    \label{fig:picture2.5} 
\end{figure*}

\vspace{3pt}
\noindent \textbf{High Physical Fidelity of the Zero-Shot PISNN.}
We first evaluated the core requirement of our framework—physical fidelity—using the one-dimensional heat equation as a canonical benchmark. Our initial information can be obtained from PINN or sensors.The resulting spatio-temporal evolution of the heat diffusion process is visually indistinguishable from the PINN reference, demonstrating exceptional fidelity Fig.~\ref{fig:picture2.5}(a,b)). This qualitative agreement is confirmed quantitatively; the point-wise absolute error remains negligible (Fig.~\ref{fig:picture2.5}(c)), and the overall Root Mean Squared Error (RMSE) across the spatio-temporal domain stabilizes at an order of magnitude of \(10^{-3}\). This provides strong evidence that our zero-shot compilation faithfully translates the continuous knowledge of the PINN into a discrete, event-driven solver without a significant loss of accuracy. 

\vspace{3pt}
\noindent \textbf{Interpretable conservation by architectural design.}
A key advantage of the PISNN lies in its mechanism for enforcing physical laws, which is both structurally guaranteed and interpretable. To demonstrate this, we tracked the total system mass during the prediction. While a well-trained PINN can be optimized to near-perfectly conserve mass, our PISNN is \textbf{conservative-by-construction}. This distinction is fundamental. The PISNN's perfect mass conservation (Fig.~\ref{fig:picture2.5}(e)) is a direct and transparent consequence of its architecture, originating from the 'flux quantization' mechanism where every spike represents a verifiable packet of physical quantity (Methods). In stark contrast, the PINN's conservation is an emergent property of its millions of optimized parameters, operating as an uninterpretable 'black box'. This structural approach establishes a more trustworthy paradigm for scientific simulation, where understanding \textit{how} a model works is as crucial as the result it produces. 

\vspace{3pt}
\begin{figure*}[t]
    \centering
    \includegraphics[width=\textwidth]{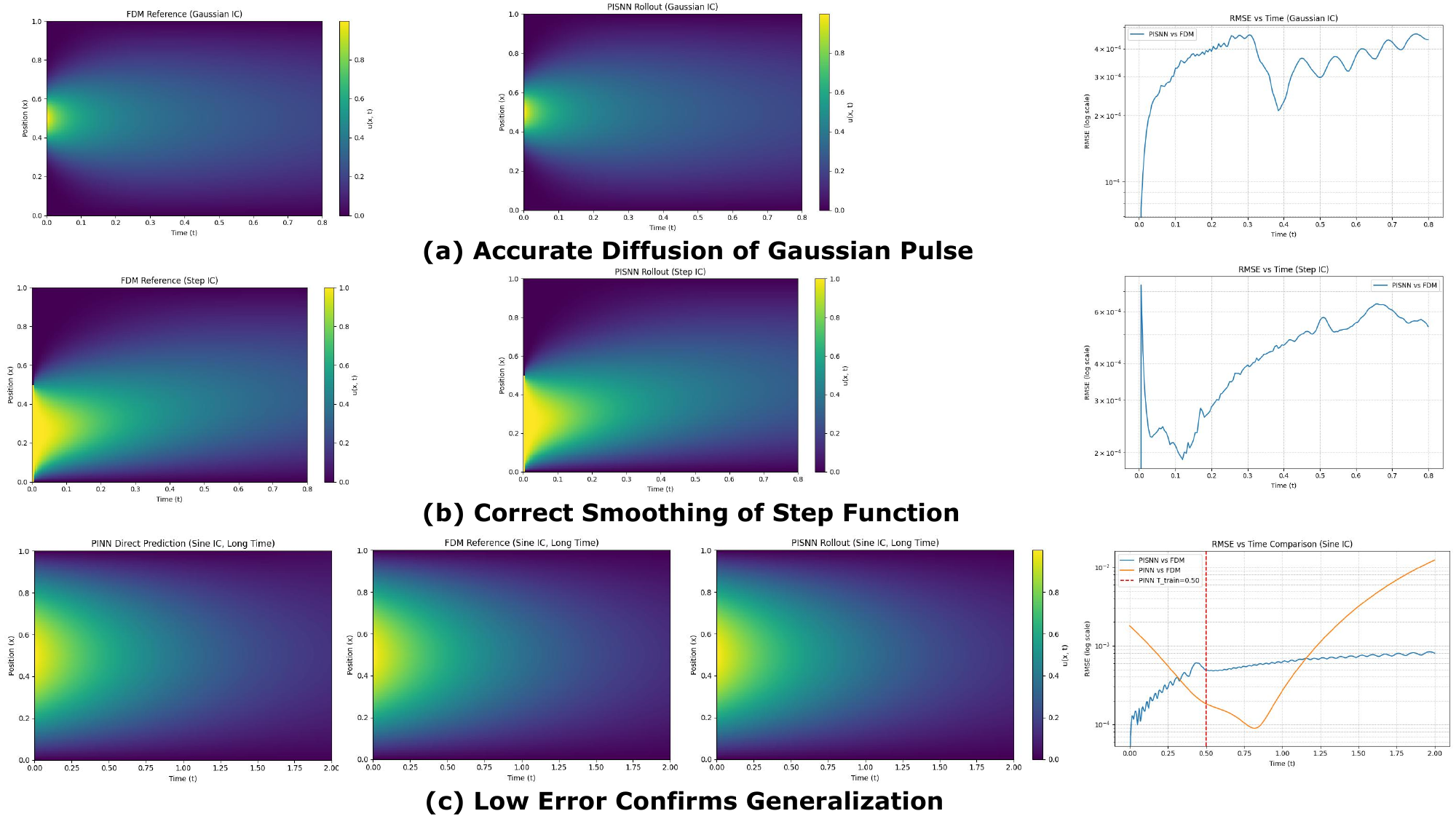}
\vspace{-10pt}
\caption{\textbf{Robust generalization in state space and time.} 
\textbf{(a-b)} Zero-shot generalization to unseen initial conditions. Even when compiled from a teacher trained only on sine waves, the PISNN accurately simulates the diffusion of a Gaussian pulse \textbf{(a)} and the smoothing of a discontinuous step function \textbf{(b)} compared to the FDM reference (Left columns). The low RMSEs (Right columns) confirm the solver's universality.
\textbf{(c)} Long-term temporal extrapolation. 
Left: FDM reference up to $T=2.0$. 
Middle: The PINN fails to generalize beyond its training horizon ($T_{train}=0.5$, dashed line). 
Right: Our PISNN maintains stability and accuracy throughout the extended simulation. The RMSE plot quantifies the PINN's divergence versus the PISNN's sustained fidelity.}
    \label{fig:picture3}
\end{figure*}

\noindent \textbf{Superior energy efficiency via event-driven sparse computation. }
A primary motivation for the PISNN framework is its potential for energy-efficient simulation on resource-constrained hardware, a known challenge for conventional PINNs. We quantitatively investigated this potential by comparing the computational cost proxies required for the PISNN and a PINN to achieve a given simulation accuracy. As illustrated in Fig.~\ref{fig:picture2.5}(f), the two paradigms exhibit a fundamental difference in their cost-accuracy profiles.

Our PISNN, by virtue of its tunable $quota$ parameter, traces a clear Pareto frontier, allowing a trade-off between accuracy and its primary computational cost proxy for neuromorphic hardware, the Total Spike Count. The conventional PINN, conversely, operates at a single point defined by its accuracy and the Total FLOPs required for inference on traditional hardware. The results reveal a stark contrast: to achieve a final RMSE comparable to or better than the PINN, the PISNN requires a Total Spike Count that is approximately three orders of magnitude lower than the PINN's Total FLOPs. While the fundamental operation for PISNN (SOP) is far more energy-efficient than the fundamental operation for PINN (FLOP). A single SOP on Loihi 2 (~0.067 pJ) is approximately 7.5 times more energy-efficient than a half-precision (FP16) FLOP on an A100 (~0.4 pJ)~\cite{NVIDIA2022A100,Intel2024HalaPoint}.This suggests that by leveraging an event-driven architecture, the PISNN paradigm can achieve high-fidelity physical simulations at a fraction of the computational budget required by dense, learning-based models.

\noindent \textbf{Robust generalization in state space and time. }
A critical test for any learning-based solver is its ability to generalize beyond its training data -- both to unseen initial conditions (state-space generalization) and to future times (temporal generalization). Our PISNN demonstrates robust performance on both fronts. First, to evaluate state-space generalization, we tested whether the PISNN internalized the underlying physical operator rather than merely overfitting to a specific trajectory. A single PISNN was compiled from a teacher PINN trained \textbf{exclusively} on a sinusoidal initial condition. This PISNN was then tasked with simulating the system's evolution from two entirely new initial states: a localized Gaussian pulse and a discontinuous step function. It accurately captures the distinct dynamics of both cases (Fig.~\ref{fig:picture3}), proving it functions as a general-purpose operator. Second, our framework overcomes a critical limitation of standard PINNs: poor long-term temporal generalization. Whereas PINNs learn a static map $u(x,t) = f_{\theta}(x,t)$ that often fails when extrapolating to times beyond the training domain, our PISNN learns a time-invariant evolution operator, $u_{t+\Delta t} = \mathcal{O}_{\theta}(u_t)$. This architectural difference allows for stable and accurate forward evolution far beyond the temporal horizon of the teacher model. This result provides compelling evidence that our framework learns the differential operator itself, functioning not as a pattern-matching network but as a true numerical solver robust to variations in both initial state and time.

\subsection*{Generalization to 2D Laplace’s Equation }
\begin{figure*}[t!]
    \centering
        \vspace{-20pt}
    \includegraphics[width=\textwidth]{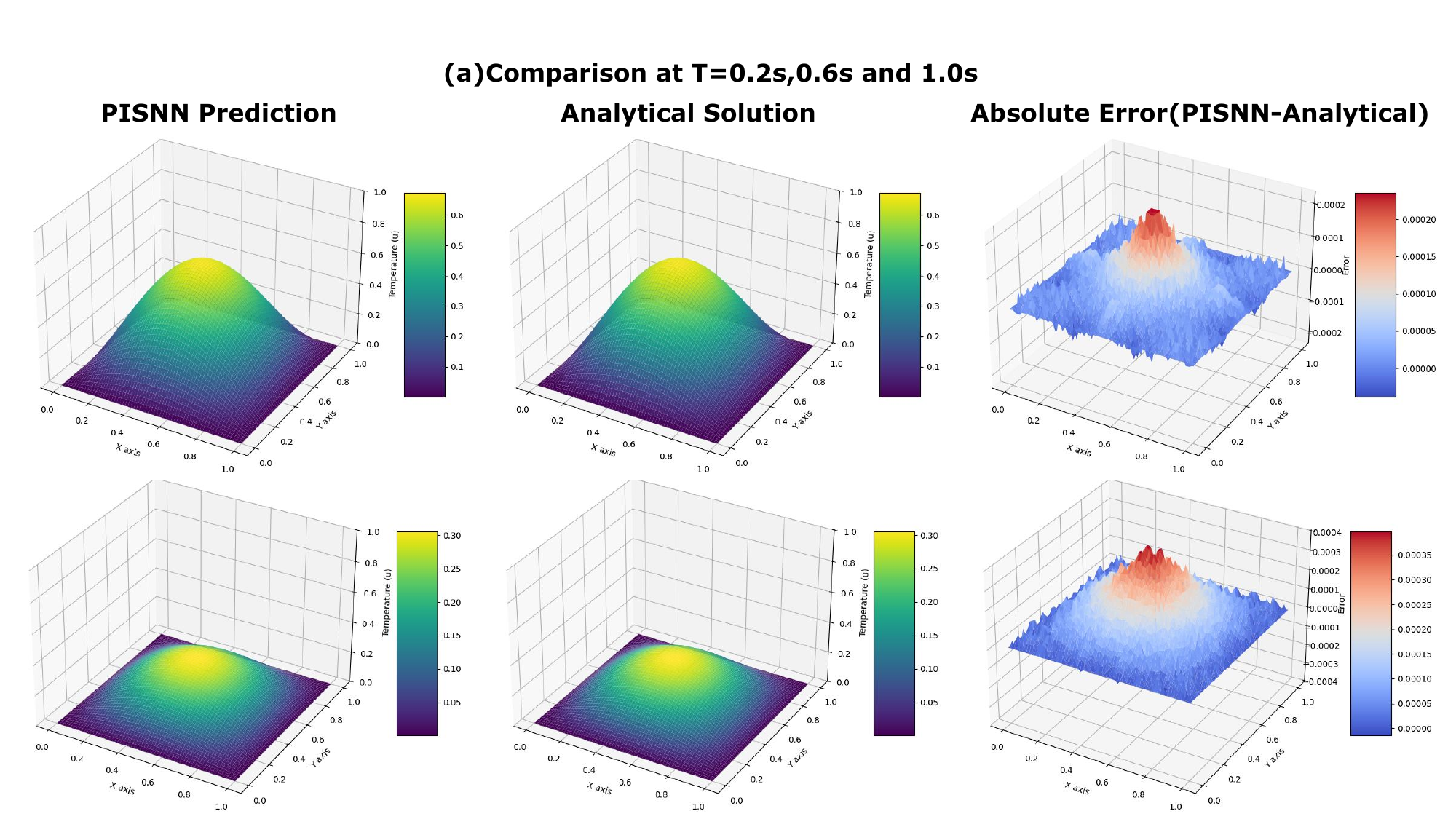}
        \vspace{-20pt}
    \includegraphics[width=\textwidth]{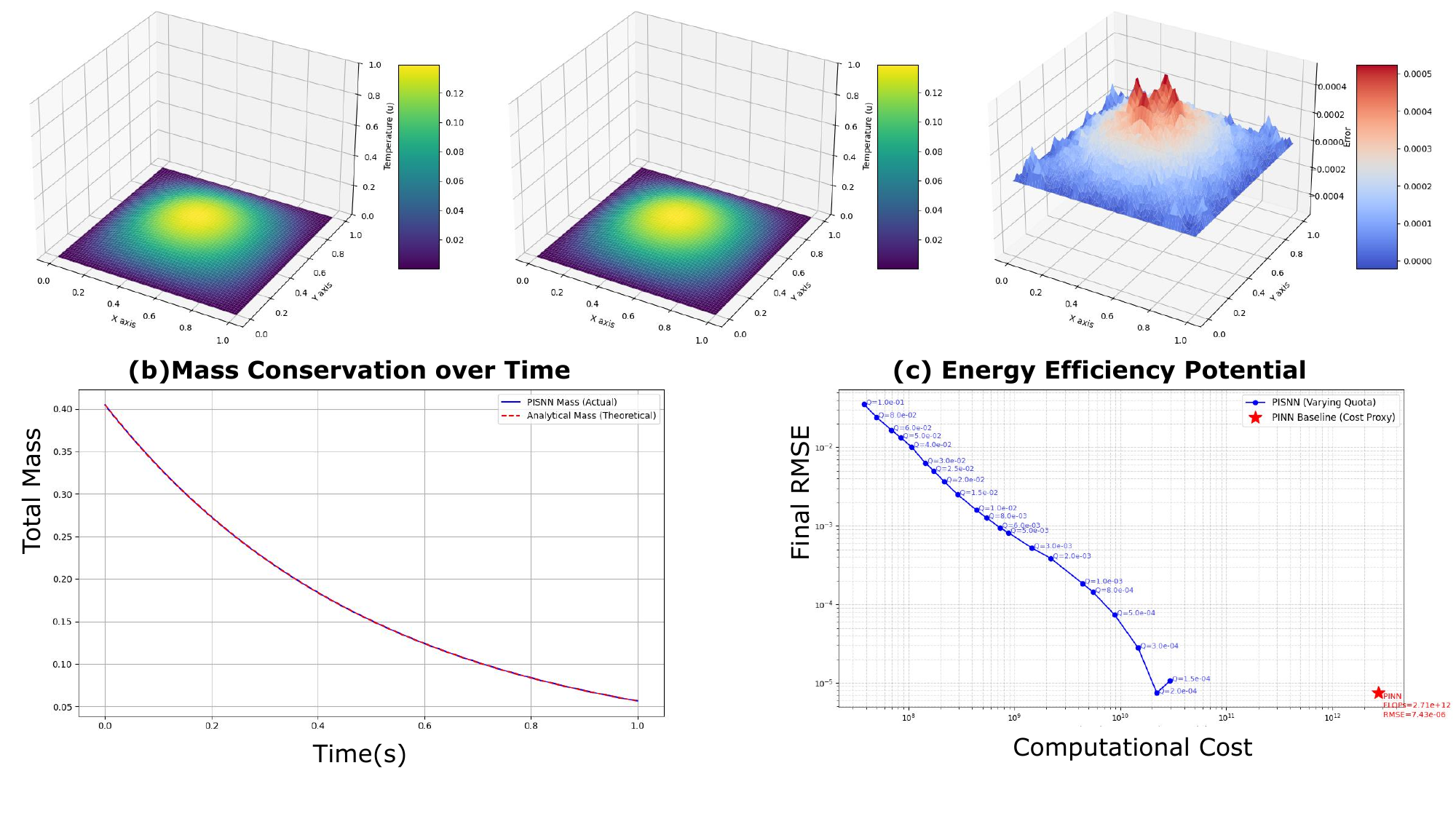}
\caption{Comprehensive validation of the PISNN on the 2D Laplace's equation. 
        \textbf{(Top rows)} Spatio-temporal comparison of the PISNN simulation (left column) against the analytical solution (middle column) at T=0.1s, 0.4s, and 0.8s. The negligible absolute error (right column) demonstrates strong functional simulation fidelity. 
        \textbf{(Bottom-left)} Total mass conservation over time. The PISNN's computed mass (purple line) perfectly matches the analytical mass (blue line), demonstrating perfect physical fidelity and strict adherence to conservation laws. 
        \textbf{(Bottom-right)} The energy efficiency Pareto frontier (Final RMSE vs. Computational Cost). The curve illustrates that the PISNN framework can achieve high accuracy (low RMSE) even at very low computational cost proxies, highlighting its significant energy efficiency potential.}
    \label{fig:picture5} 
\end{figure*}

To demonstrate that our framework's capabilities extend beyond linear diffusion, we apply the identical zero-shot compilation methodology to the 2D Laplace's equation.
Without any modification to the core framework, the PISNN successfully generalized. Visual snapshots confirm that it accurately captures the key dynamic features, remaining in close agreement with the teacher PINN. Crucially, the PISNN's two hallmark advantages persist in this more demanding scenario. First, its \textbf{conservation-by-construction} property remains perfectly intact, with total system mass conserved to machine precision (Fig.~\ref{fig:picture5}). Second, its \textbf{event-driven, sparse computation} is maintained, ensuring its significant energy efficiency advantage over the dense calculations required by a PINN is sustained even for multi-dimensional, nonlinear problems. These findings establish our PISNN framework as a versatile, robust, and efficient tool for compiling diverse physical models into accurate, inherently conservative solvers ready for edge deployment.

 \section*{Discussion}

 We introduce a physics-compilation paradigm that achieves the first complete 'spikification' of Physics-Informed Neural Networks (PINNs). Our framework reframes Spiking Neural Network (SNN) generation from an optimization problem to a zero-shot, deterministic compilation of a continuous PINN into an inherently conservative, event-driven solver. This approach simplifies deployment and embeds physical conservation directly into the network architecture.
 
\noindent \textbf{Interpretability and robustness by structural design.}
The compilation preserves the teacher PINN's high physical fidelity across linear and nonlinear systems while addressing the critical challenge of interpretability. Unlike the emergent, 'black-box' conservation of a PINN~\cite{du2025physics,mehtaj2025scientific,ranasinghe2024ginn}, the PISNN's adherence to physical laws is a transparent, structurally guaranteed consequence of its flux quantization mechanism. Each spike represents a verifiable packet of physical flux, rendering the model's behavior mechanistically verifiable and suited for high-stakes applications. The framework's robust generalization to unseen initial conditions demonstrates that it internalizes the physical evolution operator, functioning as a true numerical solver rather than a pattern-matching network. This is complemented by its computational efficiency, where the event-driven sparsity promises significant energy savings on neuromorphic hardware. The physically-principled \verb|quota| parameter provides direct, interpretable control over the accuracy-efficiency trade-off, a critical feature for edge deployment.

\vspace{3pt}
\noindent \textbf{A synergistic paradigm for neuromorphic scientific computing.}
Our work situates itself at the confluence of several research domains, creating a new synergistic paradigm. \textbf{In relation to PINNs}, our framework should be viewed not as a replacement, but as a crucial partner. While PINNs excel at learning continuous representations of physical laws from data~\cite{cuomo2022scientific}, our PISNN compiler provides the "last-mile" solution, translating these representations into efficient, discrete-event solvers. In doing so, it addresses critical PINN shortcomings by introducing structural interpretability, robust long-term generalization, and vastly lower energy overhead~\cite{krishnapriyan2021characterizing}.

This physics-centric approach also distinguishes our work from \textbf{naive PINN-to-SNN conversions}. Such methods, lacking mechanisms designed around the governing physical laws, often degrade accuracy and amplify the generalization errors of the source PINN. Our framework, by contrast, preserves high fidelity by its design. Conceptually, this represents a departure from mainstream SNN research: whereas most SNNs use spikes for efficient \textit{information coding}, our framework pioneers their use for the direct transport of \textit{physical quantities} , bridging neuromorphic engineering with scientific computing principles like the finite volume method.

Finally, when compared with \textbf{traditional numerical solvers} like FDM/FEM, the advantages are clear. While these solvers remain the gold standard for accuracy, their computational complexity makes them ill-suited for the edge~\cite{leveque2007finite}. The PISNN offers a new path, promising comparable fidelity and superior generalization capabilities with potentially orders-of-magnitude greater energy efficiency when deployed on appropriate hardware~\cite{han2018solving}.

\subsection*{Future Landscapes of PISNN}
Our PISNN framework, by compiling physical laws into an energy-efficient, brain-inspired neural models, bridges the gap between rigorous scientific computing and resource-constrained edge intelligence. This opens up potential applications for on-device physics simulation across several critical domains.

\textbf{Real-time Computational Hemodynamics for Wearable Health.} 
Current cardiovascular monitoring relies heavily on statistical correlations due to the high computational cost of solving hemodynamics (e.g., Navier-Stokes equations). Our framework offers a pathway to deploy patient-specific, high-fidelity hemodynamic models directly on wearable or bedside neuromorphic chips \cite{coorey2022health}. By solving blood flow dynamics with strict mass conservation in real-time, a PISNN-based monitor could continuously estimate localized pressure and shear stress \cite{taebi2022deep}. This enables the early detection of hemodynamic anomalies based on evolving biophysical principles rather than static snapshots, advancing the development of the cardiovascular digital twin \cite{sharon2025impact} and providing clinicians with actionable insights for preventive care.

\textbf{Onboard Physics Engines for Agile Robotics.} 
Autonomous agents, such as drones and ground vehicles, often struggle with unmodeled physical disturbances because onboard hardware cannot run complex fluid dynamics simulations in the control loop. PISNN addresses this by functioning as a low-latency "onboard physics engine" for edge computing \cite{liu2019edge}. For instance, in search-and-rescue scenarios, a drone could utilize a PISNN to estimate local thermal updrafts or turbulence via the heat equation. Similarly, for autonomous driving \cite{arafat2023vision}, PISNN could assist in real-time dynamics estimation under adverse weather conditions. This "physical intuition"—grounded in the rapid resolution of PDEs—allows control algorithms to proactively adjust to environmental dynamics, enhancing safety without relying on cloud connectivity.

\textbf{Distributed Structural Health Monitoring (SHM).} 
Traditional SHM systems typically function as passive data loggers. PISNN enables a shift towards active, predictive maintenance for critical infrastructure. By embedding neuromorphic sensors trained to solve diffusion or wave equations \cite{mehtaj2025scientific}, the structure can simulate stress propagation locally upon detecting micro-vibrations. The conservation-by-construction nature of PISNN ensures that these long-term predictions remain physically consistent, directly addressing the "black box" trust crisis associated with standard AI in safety-critical systems \cite{ribeiro2016should,ghassemi2021false,arrieta2020explainable}. This allows for precise lifetime estimation and reliable decision-making for smart infrastructure.

\subsection*{Limitations and Future Directions}

Despite its promise, our method has several limitations that define clear avenues for future research. First, while PISNN learns a time-invariant operator, like any numerical solver, it is susceptible to the long-term accumulation of discretization errors. This opens a promising direction in creating hybrid "digital twin" systems for embodied agents~\cite{mustafee2023hybrid,stupnytskyy2024digital}. The PISNN could serve as a fast, predictive forward model, periodically recalibrated by sparse, real-world data, ensuring both long-term accuracy and real-time performance.

A key challenge is to enhance the versatility of the PISNN operator itself. While we demonstrated success on structured grids, extending the flux quantization concept to complex geometries on unstructured meshes remains an open challenge, perhaps by integrating principles from Graph Neural Networks~\cite{pfaff2020learning,sanchez2020learning}. Moreover, a crucial next step is to develop end-to-end training schemes. Such schemes could be combined with a fully adaptive \verb|quota| mechanism that dynamically adjusts the quantization resolution based on local dynamics, significantly enhancing the solver's autonomy and efficiency.

The overall computational cost of our PISNN includes not only the spike communication and synaptic operations inherent to SNNs but also the calculations performed by the Flux Calculator and Spike Projector modules. 
While these steps introduce overhead compared to a purely spike-based model, they are essential for embedding the physical dynamics and ensuring conservation. Crucially, however, these auxiliary calculations can also leverage the system's sparsity; flux gradients and quantization may only need to be computed intensively in spatially active regions where significant changes occur. Therefore, we posit that the total spike count and resulting synaptic operations remain the dominant factors and key proxies for energy consumption on suitable neuromorphic hardware, assuming that these necessary arithmetic operations for flux calculation and projection can be implemented efficiently on-chip or tightly coupled to the spiking core. Realizing PISNN's full energy-saving potential thus hinges on co-designing hardware that efficiently handles both sparse, event-driven communication and these localized, physics-based computations~\cite{schuman2022opportunities,sze2017efficient}.

Future work will focus on deploying the PISNN architecture on platforms such as Intel's Loihi 2 or the SpiNNaker system to empirically quantify the energy savings predicted by our spike-based analysis~\cite{davies2018advancing,furber2014spinnaker}. Success in these areas, along with extending the framework to more complex systems, will establish this physics-compilation paradigm as a robust and indispensable tool for the next generation of intelligent autonomous systems

\section*{Methods}
This section 
presents the technical details for constructing Physics-Informed Spiking Neural Network (PISNN) solvers that simultaneously achieve high physical fidelity and robust temporal generalization. To address the challenge of physical fidelity, we introduce the Conservative Leaky Integrate-and-Fire (\textbf{C-LIF}) neuron, a novel model designed to strictly adhere to physical conservation laws at the microscopic level. To ensure temporal generality, we propose the Continuous Flux Spike Conversion (\textbf{CFSC}) method, which enables the network to accurately evolve the system over long time horizons without error accumulation. These two core components are integrated into a cohesive methodology to the underlying physics and generalize well across extended temporal domains.

\begin{table*}[t]
\begin{tabularx}{\textwidth}{@{} l l X X X @{}}
\toprule
\textbf{Mechanism} & \textbf{Post-Spike Action} & \textbf{Physical Analogy} & \textbf{Information Preservation} & \textbf{Conservation Guarantee} \\
\midrule
Hard Reset & V→Vreset & Artificial reset & High loss (state is erased) & None \\
\addlinespace
Soft Reset & V→V-Vth & Heuristic adjustment & Partial & None \\
\addlinespace
Conservative Subtraction (C-LIF) & V→V-flux\_out & Exact physical transfer & Complete (state reflects balance) & Structurally enforced \\
\bottomrule
\end{tabularx}
\caption{\textbf{Comparison of LIF neuron reset mechanisms.} The table categorizes distinct reset protocols based on their mathematical formulation and physical implications. Unlike traditional Hard and Soft resets which result in state erasure and lack conservation guarantees, the proposed C-LIF employing 'Conservative Subtraction' uniquely achieves exact physical accounting and structural conservation enforcement.}
\label{tab:picture6}
\end{table*}

\subsection*{The C-LIF Neuron}
The cornerstone of our PISNN framework is the C-LIF neuron, a physicalized adaptation of the standard LIF model. In our framework, each C-LIF neuron directly represents a discrete control volume within the physical simulation grid. Its primary state variable, the membrane potential \(V(t)\), is not an abstract activation but a direct map to a physical quantity, such as temperature, concentration, or pressure, within that volume.
The dynamics of a C-LIF neuron are governed by the following differential equation:
\begin{equation}
    \tau_m \frac{dV(t)}{dt} = - (V(t) - V_{rest}) + R_m I_{syn}(t),
\end{equation}
where each parameter possesses a distinct physical meaning:
\begin{itemize}
    \item \textbf{\(V(t)\)}: The membrane potential, directly corresponding to the physical quantity of the control volume.
    \item \textbf{\(\tau_m\)}: The membrane time constant, representing the relaxation time of the physical system. It can be related to physical timescales (e.g., \(\frac{dx^2}{\kappa}\) in heat transfer) or set to 1 for simplification.
    \item \textbf{\(V_{rest}\)}: The resting potential, corresponding to the system's background or equilibrium state (e.g., ambient temperature).
    \item \textbf{\(I_{syn}(t)\)}: The synaptic current, representing the net sum of physical quantities exchanged with adjacent neurons. It is calculated from the discrete fluxes, \(f_{\text{disc}}\), across all faces of the control volume:
    \begin{equation}
        I_{syn}(t) = \frac{1}{\Delta V} \sum_{\text{face}} f_{\text{disc,face}},
    \end{equation}
where \(\Delta V\) is the volume of the control element (e.g., \(\Delta V = dx\) in 1D).
\end{itemize}
The most critical innovation of the C-LIF neuron is its reset mechanism, which replaces the non-physical reset protocols of traditional LIF models. Standard "hard" or "soft" resets artificially destroy information and violate conservation laws by forcing the membrane potential to a predetermined value after a spike.
The C-LIF neuron completely discards this artificial reset. The evolution of its membrane potential \(V(t)\) is determined \textbf{solely and completely} by the physical fluxes represented by the synaptic current \(I_{syn}(t)\). When a spike, representing a quantum of flux, is emitted from the neuron to a neighbor, an equivalent self-inhibitory current is simultaneously applied to the neuron itself. This process, termed \textbf{`Conservative Subtraction'}, precisely deducts the outgoing physical quantity from its internal state. This is not a state reset but an exact physical accounting, which structurally guarantees local and global conservation of the physical quantity.

As illustrated in Tab. ~\ref{tab:picture6}, Conservative Subtraction is the only physically plausible mechanism that fully preserves state information and structurally guarantees conservation, making it the requisite choice for high-fidelity physical simulation.
 
\subsection*{Conservative Flux Quantization (CFQ)}

We propose \textbf{Conservative Flux Quantization (CFQ)}, a paradigm shift that discretizes continuous \textit{physical fluxes} rather than state variables directly. Rooted in the integral form of conservation laws ($\frac{\partial u}{\partial t} + \nabla \cdot \mathbf{F}(u) = S$), this "flux-first" approach ensures that every state update strictly adheres to physical constraints via the C-LIF mechanism. The framework operates through three synergistic modules:
\begin{enumerate}
    \item \textbf{Physical Flux Calculator:} Acting as the physics engine, this module maps the discrete state $u_t$ to continuous interfacial fluxes $\mathbf{F}_{\text{phys}}$ using finite difference schemes and system-specific constitutive relations (e.g., $\mathbf{F} = -\mathbf{K}(u) \nabla u$).
    \item \textbf{Spike Projector:} Serving as the quantization bridge, it converts continuous fluxes into discrete spike events via linear quantization ($N = \mathbf{F}_{\text{phys}} / \text{quota}$) and rounding. The reconstructed discrete flux, $f_{\text{disc}} = n \cdot \text{quota}$, encodes physical information into integer-based neural activity.
    \item \textbf{Discrete Flux Processor (DFP):} Functioning as the conservative solver, it evolves the system by computing the net flux divergence $(\nabla \cdot \mathbf{f})$ and updating the state via a high-order numerical integrator.
\end{enumerate}

A critical innovation of our method is that the quantization granularity, governed by the parameter \texttt{quota}, is learned rather than heuristically assigned. We treat the determination of \texttt{quota} as a teacher-student distillation problem, utilizing a pre-trained PINN as the physics oracle. The parameter is optimized by minimizing a composite loss function $\mathcal{L}_{total}$ that explicitly balances simulation fidelity against neuromorphic energy consumption:
\begin{equation}
    \mathcal{L}_{total} = \underbrace{|| u_{pred} - u_{teacher} ||^2_2}_{\text{Fidelity}} + \lambda \cdot \underbrace{\sum |s_{i,t}|}_{\text{Sparsity}}
\end{equation}
Here, the fidelity term (MSE) aligns the PISNN's trajectory with the high-precision PINN teacher, while the sparsity term (an $L_1$ regularization on spike count $s_{i,t}$) enforces energy efficiency. The hyperparameter $\lambda$ modulates the accuracy-efficiency trade-off.
Minimizing $\mathcal{L}_{total}$ yields an optimized \texttt{quota}, effectively parameterizing a discrete-time evolution operator $\mathcal{O}_\theta$:
\begin{equation}
    u_{t+\Delta t} = \mathcal{O}_\theta(u_t) = u_t - \Delta t \cdot \mathcal{D}\Big(\mathcal{Q}_{\theta}\big(\mathcal{F}(u_t)\big)\Big)
\end{equation}
Unlike standard deep learning approaches that approximate a static solution map $u(\mathbf{x}, t) = f(\mathbf{x}, t)$ and degrade rapidly during extrapolation ($t > T$), our method learns the intrinsic, time-invariant dynamical rule $\mathcal{O}_\theta$. Consequently, the learned operator supports indefinite recursive application ($u_0 \to u_1 \to \dots$), achieving \textbf{zero-shot time generalization} and \textbf{structural stability} superior to conventional non-conservative solvers.


\subsection*{PISNN Framework}

Based on CFQ, we finally can build up the PISNN framework that serves the first systematic discretization of PINNs), translating continuous physical systems into energy-efficient brain-inspired neural solvers. Constructed as a locally-connected recurrent network, its topology is rigorously isomorphic to the discretized physical domain, ensuring both computational efficiency and direct physical interpretability. Specifically, for a two-dimensional physical domain with dimensions $L_x \times L_y$ and discretization steps $d_x, d_y$, the network architecture is strictly defined such that the \textbf{Size} corresponds to $L_x/d_x$ and the \textbf{Layer} depth corresponds to $L_y/d_y$. This geometric mapping establishes a one-to-one correspondence where each C-LIF neuron represents a specific spatial control volume. Connectivity is sparse and local, with neurons establishing synaptic connections exclusively with their immediate spatial neighbors as dictated by the differential operator's stencil, thereby eliminating the redundancy of fully connected layers.

Eschewing the dense matrix operations of traditional artificial neural networks, the PISNN evolves purely through asynchronous, event-driven spike exchange. The membrane potential $V(t)$ of a neuron is updated \textbf{solely upon the arrival of discrete spikes} from neighboring units. Each incoming spike integrates a quantized packet of physical flux—defined by the \texttt{quota} parameter—into the recipient's state. Simultaneously, the emitting neuron executes 'Conservative Subtraction' to deduct the outgoing flux, ensuring structural conservation. This mechanism restricts computation exclusively to regions of active flux exchange, yielding high computational sparsity.

The deployment of this framework follows a systematic three-phase pipeline comprising Initialization, Calibration, and Real-time Correction. First, the neuron grid is instantiated based on the physical domain dimensions, with stability criteria (e.g., CFL condition) pre-computed to guarantee numerical robustness. Subsequently, the quantization parameter \texttt{quota} is optimized via a teacher-student distillation process, where a pre-trained PINN guides the minimization of a fidelity-sparsity composite loss. Crucially, the deployed PISNN is designed for dynamic interaction with the physical world; it can ingest sparse, real-time data from external sensors during inference. These observational inputs allow the solver to periodically recalibrate its internal state or fine-tune the \texttt{quota} parameter on-the-fly, effectively closing the loop between the digital solver and the evolving physical reality to prevent long-term drift.

\subsection*{Metrics for Evaluation}

To rigorously assess the performance of our PISNN, we quantify its accuracy, physical consistency, and computational efficiency using the following metrics.

\noindent \textbf{Spatiotemporal Accuracy (RMSE).} 
We evaluate the simulation fidelity by computing the Root Mean Squared Error (RMSE) between the predicted state $u_{\text{pred}}$ and the reference solution $u_{\text{ref}}$ (analytical or PINN) over the discretized 2D spatial domain $\Omega$:
\begin{equation}
    \text{RMSE}(t) = \sqrt{\frac{1}{N_{grid}} \sum_{\mathbf{x} \in \Omega} \left( u_{\text{pred}}(\mathbf{x}, t) - u_{\text{ref}}(\mathbf{x}, t) \right)^2}
\end{equation}
where $N_{grid} = N_x \times N_y$ represents the number of grid points. This metric allows tracking the accumulation of numerical errors over the simulation time $T_{eval}$.

\noindent \textbf{Operator Fidelity (Relative $L_2$ Error).}
To verify that the PISNN correctly internalized the physical evolution operator, we measure the Functional Simulation Error. This metric quantifies the relative discrepancy between a single PISNN step and the reference evolution, given an identical starting state $u(t)$:
\begin{equation}
    \text{Err}_{\text{func}}(t) = \frac{\| u_{\text{snn}}(t+\Delta t) - u_{\text{ref}}(t+\Delta t) \|_2}{\| u_{\text{ref}}(t+\Delta t) \|_2}
\end{equation}
where $\| \cdot \|_2$ denotes the Euclidean norm.

\noindent \textbf{Conservation Analysis (Total Mass).}
A key validation of our framework is its structural adherence to conservation laws. We calculate the total physical mass $M(t)$ within the domain using a discrete Riemann sum approximation, consistent with our Finite Volume formulation:
\begin{equation}
    M(t) = \Delta x \Delta y \sum_{i=1}^{N_x} \sum_{j=1}^{N_y} u(x_i, y_j, t)
\end{equation}
The evolution of $M(t)$ is monitored to detect any non-physical leakage or generation of mass, comparing the PISNN's trajectory against the theoretical value derived from the analytical solution.

\noindent \textbf{Neuromorphic Energy Efficiency.}
We assess computational cost using two proxies:
\begin{itemize}
    \item \textbf{Total Spike Count ($N_{\text{spikes}}$):} Representing the energy consumption on neuromorphic hardware, this metric sums the absolute spike events across all interfaces. For the RK4 integrator used in our DFP, the count is averaged over the four stages to represent the effective flux transfer per step:
    \begin{equation}
        \text{Spikes}^{(k)} = \frac{1}{4} \sum_{s=1}^{4} \sum_{\text{interfaces}} \left( n_{+}^{(k,s)} + n_{-}^{(k,s)} \right)
    \end{equation}
    \item \textbf{Computational Sparsity ($\bar{S}$):} Defined as the fraction of inactive interfaces per step, highlighting the event-driven nature where computation occurs only at regions of non-zero flux.
\end{itemize}
To compare with the standard PINN solvers, we also compute the Total FLOPs for the baseline PINN using the \texttt{thop} library, estimating the dense matrix operations required for an equivalent simulation duration.
\bibliography{references}

\begin{appendices}

\subsection*{Supplementary Information}
No supplementary information.
\end{appendices}

\end{document}